%% file: main.tex
\documentclass[runningheads]{llncs}
\usepackage[T1]{fontenc}
\usepackage{graphicx}
\usepackage[T1]{fontenc} 
\usepackage{subcaption}
\usepackage{multirow}

\begin{document}
\title{Uncertainty Estimation by Human Perception versus Neural Models}

\author{Pedro Mendes\inst{1,2} \and
Paolo Romano\inst{2} \and
David Garlan\inst{1}}
\institute{Software and Societal Systems Department, Carnegie Mellon University \and
INESC-ID and Instituto Superior Técnico, Universidade de Lisboa
%Springer Heidelberg, Tiergartenstr. 17, 69121 Heidelberg, Germany
}
\maketitle              % typeset the header of the contribution

\input{sec/0_abstract}

\input{sec/1_intro}

\input{sec/2_relatedWork}

\input{sec/3_study}
\input{sec/4_results}

\input{sec/5_conclusion}

\section*{Acknowledgments}

This work was supported by the Fundação para a Ciência e a Tecnologia (Portuguese Foundation for Science and Technology) through the Carnegie Mellon Portugal Program under grant SFRH/BD/151470/2021, and by projects UIDB/50021/2020,  C645008882-00000055.PRR and C628696807-00454142 (Center for Responsible AI), 101189689 (ACHILLES).
This work was developed within the scope of the project no. 62 - ``Responsible A'', financed by European Funds, namely "Recovery and Resilience Plan - Component 5: Agendas Mobilizadoras para a Inovação Empresarial", included in the NextGenerationEU funding program.
\bibliographystyle{splncs04}
\bibliography{biblio}
\end{document}

%% file: sec/0_abstract.tex
\begin{abstract}

%Estimating uncertainty is critical for both human decision-making and artificial intelligence (AI) systems. 
Modern neural networks (NNs) often achieve high predictive accuracy but are poorly calibrated, producing overconfident predictions even when wrong. This miscalibration poses serious challenges in applications where reliable uncertainty estimates are critical.
In this work, we investigate how human perceptual uncertainty compares to uncertainty estimated by NNs. Using three vision benchmarks annotated with both human disagreement and crowdsourced confidence, we assess the correlation between model-predicted uncertainty and human-perceived uncertainty. Our results show that current methods only weakly align with human intuition, with correlations varying significantly across tasks and uncertainty metrics. Notably, we find that incorporating human-derived soft labels into the training process can improve calibration without compromising accuracy. These findings reveal a persistent gap between model and human uncertainty and highlight the potential of leveraging human insights to guide the development of more trustworthy AI systems.

%This study investigates the relationship between human perceptual uncertainty and the uncertainty estimates generated by neural networks (NNs). We evaluate the correlation between human and model uncertainty using several popular models and benchmarks across tasks with varying levels of complexity and analyze whether task complexity influences this relationship.

%The results show no significant correlation between human and model uncertainty estimates, even when accounting for input complexity. Thus, next, we investigated the incorporation of human insights into model uncertainty evaluations, which improves the correlation, suggesting that current methods are insufficient to bridge the gap between human and model uncertainty reasoning.

%These findings highlight fundamental differences in how humans and models estimate uncertainty, underscoring the need for better uncertainty estimation techniques. By improving the correlation between model and human uncertainty, we can enhance the trustworthiness and practical applicability of AI systems.
\end{abstract}

%% file: sec/1_intro.tex
\section{Introduction}
\label{sec:intro}

Neural networks (NNs) have achieved remarkable success across a variety of tasks, from image classification to medical diagnosis. However, despite their high predictive accuracy, these models often suffer from poor calibration; their confidence scores do not reliably reflect the likelihood of correctness. This issue becomes especially problematic in high-stakes applications, where overconfident yet incorrect predictions can lead to severe consequences. Due to their black-box nature, it is far from trivial to understand and explain the outputs produced by modern large and complex NNs~\cite{xai,xai_unc}.

Uncertainty estimation has emerged as a key component in building more trustworthy AI systems. Several methods~\cite{cals,deup,isotonic_regression,ce_pe,ece,euat,focal_loss,focal_loss1,dun,beta_calib,acc_unc_calib,bin_free_calib,soft_calibration} have been proposed to improve model calibration, including temperature scaling, Bayesian approximations, and ensemble techniques. While these approaches often improve alignment between predicted probabilities and observed outcomes, it remains unclear whether the resulting uncertainty estimates align with human-perceived uncertainty.

Further, real-world applications, such as medical diagnostics, autonomous driving, and fraud detection, increasingly leverage model uncertainty to enhance trustworthiness by incorporating human oversight. In these systems, models can flag predictions with high uncertainty as requiring human review. This approach allows systems to balance automation and human judgment, ensuring that critical decisions are not solely reliant on potentially unreliable predictions. By raising alarms for ambiguous cases, these systems create a feedback loop where humans can validate, correct, or refine predictions, thus improving the model's performance and fostering user trust. %Such collaborative frameworks emphasize transparency and accountability, addressing ethical concerns while enhancing the reliability of AI-driven decisions in high-stakes domains.
However, the deployment of uncertainty-based alarm systems creates a trade-off between the frequency of alarms raised and the associated costs. These costs can be reduced by aligning the model uncertainty estimation with human perception.

In this work, we explore the following question: 
To what extent do uncertainty estimates from modern NNs reflect the uncertainty perceived by humans?
If models and humans perceive uncertainty differently, then even well-calibrated models may fail to support downstream decisions in a human-compatible manner.

To investigate this, we study three visual classification tasks that include human annotations of perceptual uncertainty, either through soft labels (reflecting inter-annotator disagreement) or explicit confidence scores. We compare these human-derived signals with uncertainty estimates from multiple NNs using metrics such as predictive entropy (PE)~\cite{pe_mi,pe_mi1}. We further test whether injecting human soft labels into training can improve the model’s alignment with human intuition and its calibration.

Our key contributions are as follows:
\begin{itemize}
    \item We present a systematic comparison between model-estimated and human-perceived uncertainty across three vision benchmarks.
    \item We quantify the alignment between human and model uncertainty and analyze how it varies across datasets and uncertainty estimation techniques.
    \item We show that integrating human soft labels into training can improve model calibration without degrading accuracy, indicating the potential of human-informed training.
    \item Our findings suggest that model confidence alone may not be sufficient for trustworthy human-AI interaction and motivate future work toward hybrid uncertainty estimation approaches.
\end{itemize}

%% file: sec/2_relatedWork.tex
\section{Related Work}
\label{sec:rw}

A significant body of work has explored methods to enhance calibration of NNs.
Early methods leveraged Bayesian Neural Networks (BNNs)~\cite{BayesNN,BayesNN1,BayesNN2,BNN}, which place a prior distribution over the model weights and infer a posterior given the data~\cite{epis_alea_unc}. While theoretically sound, BNNs are computationally intensive and scale poorly to large models and datasets. To mitigate this, several approximate Bayesian methods have been proposed such as Monte Carlo (MC) Dropout~\cite{mcdropout1} and Variational Inference (VI)~\cite{Var_inf}. 
Among these, MC Dropout, due to its higher efficiency, has become the most widely used technique for uncertainty estimation in NNs~\cite{euat,uncertainty_survey,mcdropout1}. This technique estimates uncertainty by applying multiple dropout masks and performing multiple stochastic forward passes, effectively simulating an ensemble of subnetworks. The resulting predictions are aggregated using statistical measures, such as PE or variance, to quantify the model’s uncertainty.

In addition to uncertainty quantification, calibration has become a key property of NNs, as they are often overconfident when their predictions are incorrect~\cite{model_calibration}. To address this, a range of post-hoc calibration methods have been developed, including Platt Scaling~\cite{platt}, Isotonic Regression~\cite{isotonic_regression}, Temperature Scaling~\cite{model_calibration}, and Beta Calibration~\cite{beta_calib}, which adjust model confidence scores after training. These techniques are computationally efficient and easy to apply, but do not alter the underlying uncertainty estimates. Consequently, their effectiveness can vary significantly across architectures and datasets~\cite{cals}, and they tend to degrade under distributional shifts, reducing reliability in out-of-distribution (OOD) scenarios~\cite{calib_ood,model_calibration}.

Beyond post-hoc methods, a growing body of work seeks to incorporate uncertainty estimation directly into the training process. These uncertainty-aware training methods aim to jointly optimize for accuracy and calibration. Approaches in this direction include augmenting standard loss functions with uncertainty-regularizing terms, such as Focal Loss~\cite{focal_loss} and Label Smoothing~\cite{focal_loss1}. Further, Accuracy versus Uncertainty Calibration (AvUC) loss~\cite{acc_unc_calib} explicitly balances predictive performance with calibration, integrating temperature scaling into the optimization objective.
Other advancements propose differentiable calibration metrics, such as Soft Calibration Error~\cite{soft_calibration}, which relaxes traditional binning procedures, and binning-free calibration methods~\cite{bin_free_calib}, which avoid discretization entirely. Additional efforts incorporate conformal prediction frameworks, such as the uncertainty-aware conformal loss~\cite{Uncertainty_loss1}, to better align model confidence intervals with observed outcomes. Shamsi et al.~\cite{ce_pe} proposed a composite loss function that accounts for both PE and Expected Calibration Error (ECE).
More recently, CALS~\cite{cals} introduced a class-wise uncertainty weighting scheme that emphasizes harder or more uncertain examples during training. EUAT~\cite{euat} proposed a dual-loss strategy that differentiates between correct and incorrect predictions, promoting high uncertainty on mispredictions while maintaining low uncertainty for correct ones. At last, CLUE~\cite{clue} proposes a general framework for model calibration that explicitly aligns predicted uncertainty with observed error during training.
%Despite their effectiveness, these training-based approaches are largely restricted to classification models and do not directly address the alignment between model and human-perceived uncertainty
While these methods improve statistical calibration, they do not necessarily reflect how humans perceive uncertainty.

Humans intuitively account for ambiguity and express caution under uncertain conditions, suggesting a natural way to navigate complex, unfamiliar, or ambiguous inputs. They evaluate uncertainty in a highly context-dependent manner, based on heuristics, drawing on both perceptual and contextual indications, subjective perceptions, and prior knowledge to estimate confidence levels~\cite{human_unc}. Unlike NNs, which only rely on patterns learned from data, humans integrate multiple sources of information, such as experience, visual or sensory ambiguity, and situational perspectives, to assess the certainty of their judgments~\cite{human_unc2,human_unc1}. This flexible and adaptive uncertainty evaluation allows humans to balance caution and decisiveness in uncertain situations, adjusting their responses based on the perceived risk and context.

Human-perceived uncertainty has been studied through crowdsourced confidence annotations and disagreement among annotators~\cite{cifar10-h,imageNet16-h,cifar10-n}. Such signals offer a richer view of task difficulty and ambiguity. Recent work has leveraged these signals to generate soft labels for training~\cite{reidsma2008exploiting}, which can improve robustness and calibration~\cite{cifar10-h}. 
Further, Peterson et al.~\cite{cifar10-h} have shown that human labels improve generalization, i.e., the quality of the models increases when trained using soft labels obtained by human annotators. However, this work never evaluates the model uncertainty, fundamental for accessing trustworthiness. Additionally, Steyvers et al.~\cite{imageNet16-h}  developed a Bayesian modeling framework that jointly combines human and models predictions.
However, the alignment between these human-derived uncertainty signals and model-predicted uncertainty remains underexplored.

Unlike prior work, we perform a systematic, cross-dataset comparison of model-estimated uncertainty and human-perceived uncertainty. We also investigate whether incorporating soft human labels during training can improve this alignment and enhance model calibration, providing insights into the potential for hybrid human-AI trust pipelines.

%% file: sec/3_study.tex
\section{Human and Model Uncertainty Evaluation}
\label{sec:Methods}

%Study Design

Understanding and quantifying uncertainty is crucial to make informed decisions and to develop trustworthy models.
By comparing the uncertainty estimation of NNs with human perceptual uncertainty, it is possible to understand how well current methods align with human uncertainty, shedding light on both the strengths and limitations of current uncertainty estimation techniques, leading to the design of improved methods that enhance model trustworthiness.

Therefore, this work conducts a correlational study to compare human perception of uncertainty against uncertainty estimation outputted by NNs. The objectives are twofold: i) to evaluate how well model uncertainty aligns with human perceptual uncertainty, especially in ambiguous or uncertain scenarios, and ii) to explore whether model uncertainty can be enhanced or complemented by human decision-making. By identifying scenarios where human and model uncertainty assessments converge, diverge, or reveal poor calibration, we can tailor the training to develop more trustworthy models.

This study hypothesizes that human perceptual uncertainty strongly correlates with the uncertainty estimates generated by NNs. Specifically, we posit that as human uncertainty increases, model uncertainty should also increase. Furthermore, we hypothesize that the strength of this correlation might be dependent on task complexity, with correlations weakening as task complexity increases. 
Additionally, we propose incorporating human insights to enhance the quality of model uncertainty evaluations, thereby advancing the development of more trustworthy AI systems.
%The findings from this research will contribute to a broader understanding of model uncertainty estimation and open pathways toward more trustworthy AI systems.

%This study aims to compare human perceptual uncertainty with the uncertainty estimates generated by NNs across a set of tasks with varying levels of complexity. A correlation-based approach is employed to quantify the relationship between human and model uncertainties. %, with additional analyses to evaluate the impact of ambiguity, noise, and task complexity.

 %dataset
\subsection{Experimental Setup, Benchmarks, and Baselines}
\label{sec:expe}

\subsubsection{Datasets and Models.}
This work exploits publicly available datasets (namely, Cifar10-H~\cite{cifar10-h}, CifarN~\cite{cifar10-n}, and ImageNet-16H~\cite{imageNet16-h}) in the image recognition domain (normally used to train NNs) that include human annotations from multiple reviewers. The datasets containing human annotations have been labeled by multiple reviewers, with the number of reviews per image varying based on the benchmark used~\footnote{ImageNet-16H and Cifar-N dataset received approximately four to six judgments per image, while Cifar-10H  dataset has fifty annotations per image.}.  
%For instance, the Cifar-10H dataset received approximately fifty judgments per image, while the Cifar-N dataset has only four annotations per image. 
Further, ImageNet-16H extends the ImageNet~\cite{imagenet} dataset by reducing the sample size to 4800 and limiting the number of classes to 16, while introducing noise in the images. 
%Each image in this dataset receives an average of six human classifications.

%models
We considered pre-trained models, such as ResNet50~\cite{resnet} using ImageNet, as well as models trained specifically on the benchmarks used in this study, including ResNet18 on CIFAR10-H and CIFAR-N. For these benchmarks, we randomly split the data into training and testing sets, using an 80/20 ratio, respectively. Additionally, for ResNet50 on ImageNet-16H, we fine-tuned the model using the original dataset inputs and evaluated it on the noisy data containing human annotations.

To train the models, independently of the considered solution, we use stochastic gradient descent to minimize the Cross Entropy (CE) loss function using a learning rate of 0.1, a momentum of 0.9, and a batch size of 64 for all the models. 
To guarantee reproducibility, ensure a fair comparison, and mitigate the randomness inherent in training NNs, each method was trained using ten different seeds~\cite{deep_ensembles}. Throughout this study, when a subset of the dataset is required, stratified sampling techniques are applied to ensure that all classes are adequately represented and that the sample remains representative of the original dataset. Further, the models are evaluated on the same test set used for human annotations to ensure a consistent comparison between human and model uncertainty estimates.
All models and training procedures were implemented in Python3 via the Pytorch framework and trained using a single Nvidia RTX A4000 GPU.

 %Model Uncertainty Estimation
\subsubsection{Baselines.}
Furthermore, we resort to several state-of-the-art methods \cite{deup,cals,euat,deep_ensembles,model_calibration} that have been proposed to estimate model uncertainty. Moreover, one important foundation of all these works lies in the computation of uncertainty given a prediction. One of the most widely used techniques is MC Dropout \cite{MCdropout}, which provides a Bayesian approximation for uncertainty by sampling multiple dropout masks and aggregating the predictions using PE. Like the human uncertainty estimation, MC Dropout combined with PE does not separate epistemic from aleatoric uncertainty, so both human and model uncertainty will reflect the total uncertainty of a prediction.
In this study, we compare different baselines to compute the model uncertainty. More in detail, we evaluate both post-processing calibration methods (resorting to Isotonic Regression~\cite{isotonic_regression}, or  DEUP~\cite{deup}) and uncertainty-aware training algorithms (namely, CE~\cite{ce}, EUAT~\cite{euat}, CALS~\cite{cals}, or deep ensemble~\cite{deep_ensembles}).

% Human Uncertainty
The human annotations contained on the benchmarks used can be exploited to create soft labels, representing the probability of each class given an input. From these soft labels, a distribution can be generated, allowing the computation of uncertainty, for example, using the PE. The datasets primarily consist of images of everyday objects (e.g., cars, airplanes), and thus epistemic uncertainty (stemming from a lack of knowledge) of human predictions can be assumed to be low. In this context, human uncertainty likely arises from noise, ambiguity, or inherent randomness in the data, i.e. the aleatoric uncertainty. However, the proposed method for evaluating uncertainty using human annotations does not differentiate between epistemic and aleatoric uncertainty; it quantifies only the total uncertainty.

% Process and Task Complexity Evaluation
\subsubsection{Evaluation Metrics.}
We resort to the PE of the outputted distribution to compute the respective uncertainty. By applying the same uncertainty estimation method for both human and model assessments, a direct comparison of the results becomes possible.

% Correlation Analysis
To assess the relationship between human and model uncertainties, Pearson’s correlation coefficient is computed for each baseline. Additionally, we compute the correlations for each subgroup composed of inputs with differing complexity. Statistical significance was assessed at a threshold of $p<0.05$.

\subsection{Methods}

This study has two independent variables: uncertainty estimation method (human-derived versus model uncertainty estimates) and task complexity (easy, medium, difficult). Each group (humans and models) classified the same set of images to generate the uncertainty estimates.

\setlength{\tabcolsep}{6pt}
\begin{table*}[t]
    \centering
    \caption{%Task complexity based on human and model predictions. For a given input of class A, the prediction is correct if it is class A, or incorrect if it is any other class !A.
    Task complexity based on human and model predictions. For each input belonging to class $A$, the prediction is considered \textit{correct} if classified as $A$, and \textit{incorrect} if classified as any other class ($\lnot A$).}
    \label{tab:complexity}
    \vspace{2mm}
    
    \begin{tabular}{c|c|c|c}
    \textbf{True Class} & \textbf{Human annotation} & \textbf{Model prediction} & \textbf{Task complexity}\\ \hline
    $A$ & $A$ & $A$ & Easy  \\
    $A$ & $A$ & $\lnot A$ & Medium \\
    $A$ & $\lnot A$ & A &Difficult \\
    $A$ & $\lnot A$ & $\lnot A$ 	& Difficult \\
    \end{tabular}
\end{table*}

Moreover, the task complexity (i.e., the complexity/difficulty of classifying an image) is defined based on whether the human and model predictions match the true class. To automate the assessment of the task complexity, we compare the true class of each image with the human reviews (which can be averaged to determine a final class) and model predictions~\footnote{It should be noticed that the uncertainty is not used to evaluate the task complexity but only the predictions.}. When discrepancies arise between the true class and the predictions from humans or models, the complexity of the input is determined according to the criteria outlined in Table~\ref{tab:complexity}. 

%Human Insight Integration
At last, to explore the potential of human insights to improve NNs uncertainty estimates, a subset of human annotations was incorporated into the model's training using soft labels while computing the CE loss. The quality of the updated model’s uncertainty estimates was then re-evaluated against the original human uncertainty annotations.

%Ethical Considerations
%The study was conducted following ethical guidelines, with informed consent obtained from all participants. Data privacy was maintained by anonymizing participant responses and ensuring secure data storage.

%% file: sec/4_results.tex
\section{Results}
\label{sec:results}

This section reports the results obtained in this correlation study. %We start by evaluating the correlation between human and model uncertainty estimation. 
Table~\ref{tab:correlation} shows the correlation between human perceptual uncertainty and model uncertainty evaluated using the different techniques considered (namely, CE with MC dropout, EUAT, CALS, deep ensemble, isotonic regression, and DEUP) training a  ResNet18 on Cifar10-H and CifarN, and a ResNet50 on ImageNet-16H. In these experiments, human reviews are not incorporated into the training process.

\setlength{\tabcolsep}{12pt}
\begin{table*}[t]
    \centering
    \caption{Pearson correlation coefficient between model and human uncertainty using different baselines, models, and benchmarks.}
    \label{tab:correlation}
    \vspace{2mm}
    
    \begin{tabular}{c|c |c |c }
    \textbf{Baseline} & \textbf{Cifar10-H}&   \textbf{CifarN}  & \textbf{ImageNet-16H} \\\hline
    CE & 0.22 & 2.59E-3  & 0.34  \\
    EUAT & 0.21  & 2.60E-3  & 0.30   \\
    Ensemble & 0.21   & 5.07E-3  & 0.35   \\
    CALS  & 0.22   & 4.60E-3  & 0.32  \\
    Iso. Reg. & 0.24 & 4.20E-3  & 0.35  \\
    DEUP & 0.06 & 3.90E-3  & 0.02  
    \end{tabular}
\end{table*}

The results show no significant correlation between human and model uncertainty.
The Pearson correlation coefficient between the model and human uncertainty is up to 0.24, 5.07x$10^{-3}$, and 0.35 for ResNet18 trained on Cifar10-H and CifarN, and ResNet50 on ImageNet-16H, respectively. 
The corresponding p-values for ResNet18 on CIFAR10-H and ResNet50 on ImageNet-16H are below the significance threshold of 0.05, confirming the statistical validity of these results. However, for ResNet18 with CIFAR-N, the p-value exceeds 0.05, indicating no statistically significant correlation.
In conclusion, while statistically significant but weak correlations were observed for ResNet18 on CIFAR10-H and ResNet50 on ImageNet-16H, the analysis found no significant correlation between human perceptual uncertainty and model uncertainty estimates for ResNet18 on CIFAR-N. This suggests that there is no correlation between human perception uncertainty and model-derived uncertainty estimates.

\begin{figure*}[t]
\centering
\captionsetup{justification=centering}
    \begin{subfigure}[h]{0.45\textwidth}
    \centering
        \includegraphics[height=0.5\textwidth]{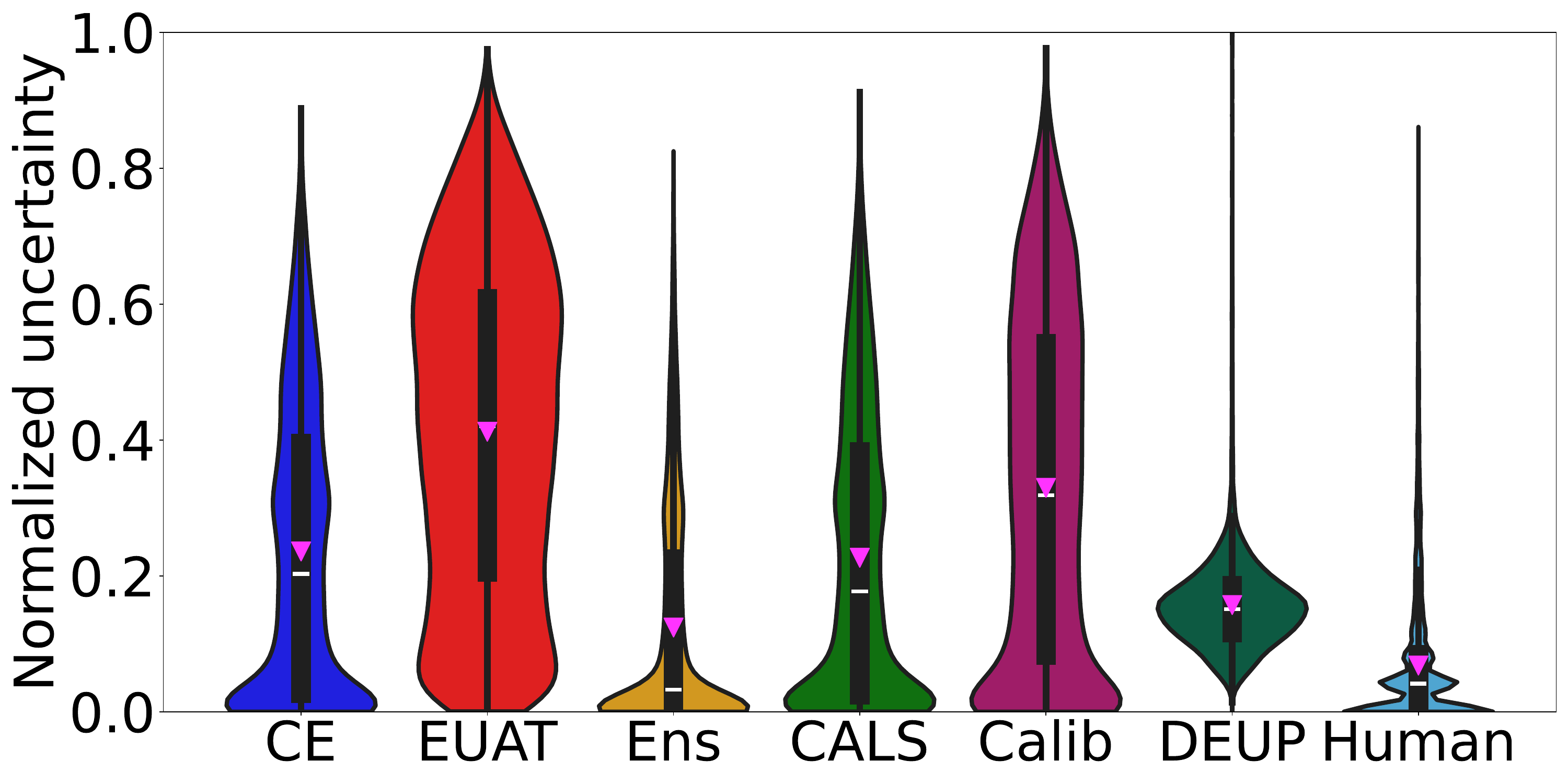}
        \caption{ResNet18 Cifar10H}
        \label{fig:unc_cifar10h}
    \end{subfigure}
    \begin{subfigure}[h]{0.45\textwidth}
    \centering
        \includegraphics[height=0.5\textwidth]{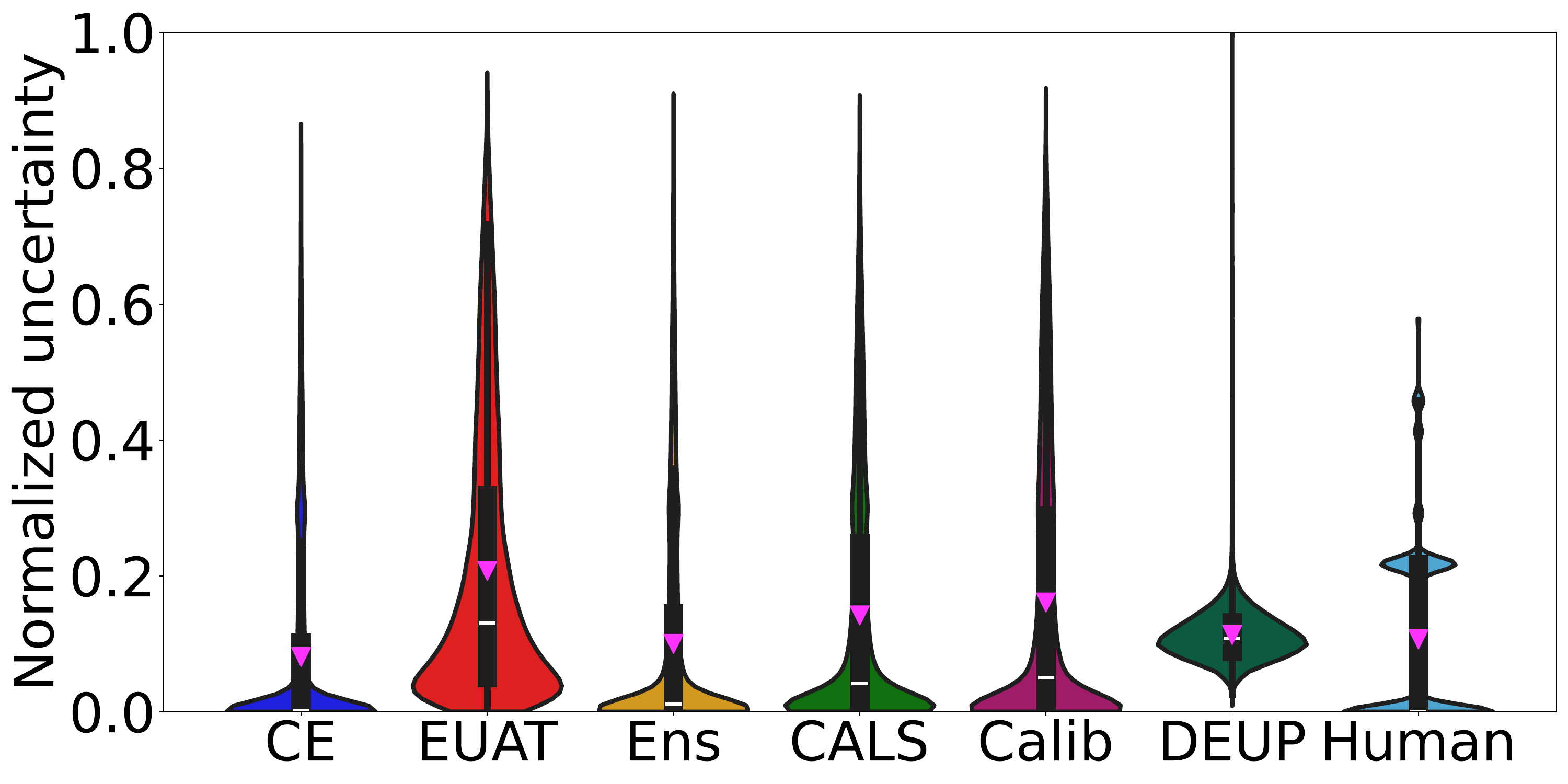}
        \caption{ResNet18 CifarN}
        \label{fig:unc_cifarn}
    \end{subfigure}
    \begin{subfigure}[h]{0.45\textwidth}
    \centering
        \includegraphics[height=0.5\textwidth]{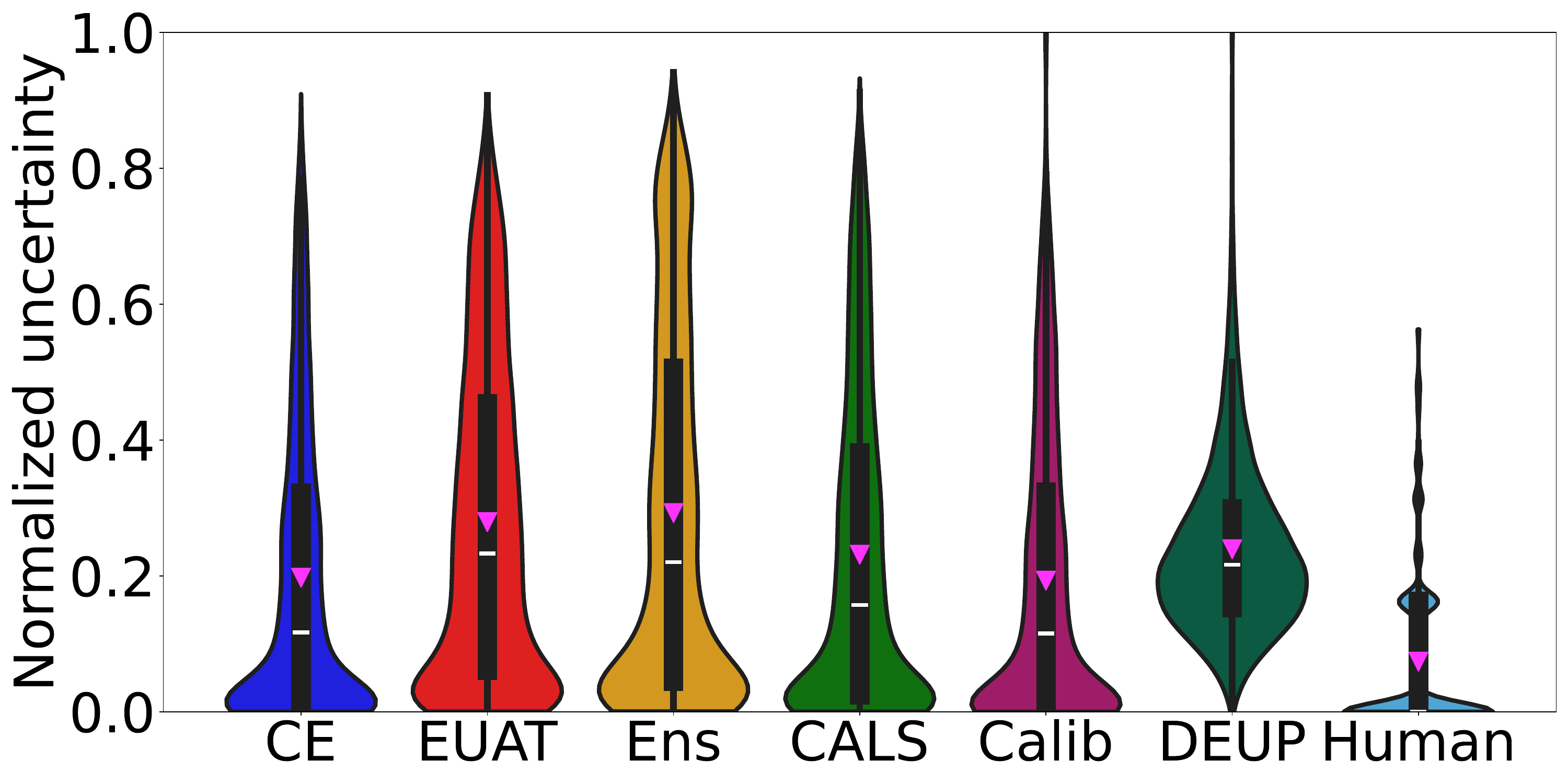}
        \caption{ResNet50 ImageNet-16H}
        \label{fig:unc_imgNet}
    \end{subfigure}
    \captionsetup{justification=justified}
    \caption{Normalized uncertainty distribution using different baselines, models, and benchmarks. (the average value of each distribution is marked with a pink triangle).}
     \label{fig:correlation}
\end{figure*}

Further, Figure~\ref{fig:correlation} plots the distribution of the normalized uncertainty of the models using several different uncertainty estimation techniques and the human uncertainty. Through the analysis of those, we see very different distributions, which visually reinforces the previous results. 

Next, we evaluate the relationship between task complexity and the correlation of uncertainty estimates by reporting, in Table~\ref{tab:complexity_results}, the Pearson coefficient of correlation between humans and multiple uncertainty estimation techniques for the different benchmarks. Once again, the results show no correlation between the task complexity and the uncertainty of humans vs. models. 

\setlength{\tabcolsep}{6pt}
\begin{table*}[t]
    \centering
    \caption{Pearson correlation coefficient between model and human uncertainty using different task complexity, baselines, models, and benchmarks.}
    \label{tab:complexity_results}
    \vspace{2mm}
    
    \begin{tabular}{c|c|c|c|c}
    \textbf{Baseline} &\textbf{Complexity} & \textbf{Cifar10-H} &   \textbf{CifarN}  & \textbf{ImageNet-16H} \\\hline
    \multirow{3}{*}{CE} & Easy & 0.21  & 0.01 & 0.22 \\
     & Medium & 0.03  & 0.0   & 0.08 \\
     & Difficult & 0.24  & 0.0   & 0.29  \\ \hline
    \multirow{3}{*}{EUAT} & Easy & 0.21  & 0.0 & 0.17  \\
     & Medium & 0.0& 0.0  & 0.05  \\
     & Difficult & 0.20  &  0.0 & 0.3  \\ \hline
    \multirow{3}{*}{Ensemble} & Easy & 0.19 & 0.02   & 0.21 \\
     & Medium & 0.02 & 0.04   & 0.01 \\
     & Difficult & 0.22 &  0.0 & 0.19  \\ \hline  
    \multirow{3}{*}{CALS} & Easy & 0.22 & 0.02   & 0.19 \\
     & Medium & 0.01  & 0.03  & 0.06  \\
     & Difficult & 0.26  & 0.0 & 0.27  \\ \hline     
    \multirow{3}{*}{Calibration} & Easy & 0.22  & 0.01  & 0.22  \\
     & Medium & 0.05 & 0.02  & 0.10  \\
     & Difficult & 0.32  & 0.0  & 0.23  \\ \hline 
    \multirow{3}{*}{DEUP} & Easy & 0.03  & 0.02   & 0.06 \\
     & Medium & 0.0  & 0.03  & 0.01  \\
     & Difficult & 0.09 & 0.01  & 0.04 \\ \hline   
    \end{tabular}
\end{table*}

While both humans and models exhibited higher uncertainty for ambiguous or noisy inputs, their uncertainty levels did not consistently align across the dataset. Furthermore, no clear pattern emerged when evaluating the relationship between task complexity and the correlation of uncertainty estimates. Although both humans and models showed increased uncertainty for more complex inputs, their responses remained uncorrelated, suggesting that task complexity does not significantly impact the alignment between human and model uncertainty. These results highlight a fundamental divergence between how humans and models assess uncertainty, even under varying task conditions.

\begin{table*}[t]
    \centering
    \caption{Pearson correlation coefficient using CE with soft labels.}
    \label{tab:correlation2}
    \vspace{2mm}
    
    \begin{tabular}{c|c|c|c}
    \textbf{Baseline} & \textbf{Cifar10-H} &   \textbf{CifarN}  & \textbf{ImageNet-16H} \\\hline
    CE with soft labels & 0.34 &  0.09  & 0.47 \\
    \end{tabular}
\end{table*}

\section{Discussion}
\label{sec:discussion}

In this study, we analyzed the relationship between human perceptual uncertainty and the uncertainty estimates generated by NNs. Contrary to our initial hypothesis, the results reveal no clear correlation between human and model uncertainty. While there are occasional instances of agreement, the overall lack of alignment suggests that models and humans rely on fundamentally different mechanisms to assess uncertainty. This misalignment raises important questions about the interpretability and reliability of model uncertainty estimates, particularly in tasks where human intuition plays a critical role.  

We also analyzed whether task complexity influences the correlation between human and model uncertainty. Once again, the results do not support a consistent relationship. While we observed that both humans and models tend to exhibit higher uncertainty for more complex inputs, their responses remain largely uncorrelated across different levels of complexity. This finding indicates that neural networks struggle to capture the nuanced patterns of uncertainty that humans intuitively recognize, even when task complexity is varied. Such discrepancies highlight limitations in current uncertainty estimation methods, particularly in tasks with ambiguous or noisy data where human expertise is crucial.  

Given these results, we further evaluated whether incorporating human insights could improve model uncertainty evaluations. Thus, a subset of human annotations was incorporated into the model using soft labels while computing the CE loss, and the correlation of the updated model’s uncertainty estimates was then re-evaluated against the human uncertainty. The results, reported in Table~\ref{tab:correlation2} show an improvement in the correlation of the model uncertainty compared to human uncertainty. More in detail, the correlation improves to 0.34 and 0.47 when training a ResNet18 with Cifar10-H and a ResNet50 with ImageNet-16H, respectively. On ResNet18 with CifarN, while there is some improvement, the correlation remains insignificant.

Further, while human annotations occasionally helped refine model predictions, their impact was inconsistent and depended on the specific nature of the task and the quality of the model’s baseline estimates. These findings suggest that simply integrating human insights may not be sufficient to bridge the gap between human and model uncertainty assessments. 
Instead, more sophisticated approaches for uncertainty estimation or designing models that explicitly account for human-like patterns of uncertainty may be necessary to improve alignment.

The lack of correlation between human and model uncertainty has important implications for the development and application of AI systems. It highlights the need for alternative methods that better emulate human uncertainty reasoning, particularly in high-stakes applications such as healthcare or autonomous driving. Without this alignment, model uncertainty estimates may lack the interpretability needed to build trust and reliability among users. These findings underscore the importance of not only improving model performance but also ensuring that uncertainty estimates reflect human-like reasoning, which is essential for societal acceptance and effective integration of AI systems into decision-making processes.

%% file: sec/5_conclusion.tex
\section{Conclusion}
\label{sec:conclusion}

This study explored the relationship between human perceptual uncertainty and the uncertainty estimates generated by NNs. Our findings indicate no significant correlation between the two, highlighting a fundamental divergence in how humans and models assess uncertainty. This behavior persisted across tasks with varying levels of complexity. While incorporating human insights into model uncertainty evaluations showed some improvements, the results emphasize the need for more robust methods that better emulate human uncertainty assessment.